%% file: maincameraready.tex
\documentclass[twoside]{article}
\usepackage[accepted]{aistats2018}

\newif\ifcomments
\commentsfalse

\usepackage{tabularx}
\usepackage{graphicx} 
\usepackage{arydshln}
\usepackage{subfigure}
\usepackage{natbib}
\usepackage{algorithm}
\usepackage{algorithmic}
\usepackage{aliascnt}
\usepackage{hyperref}
\usepackage{bbm}
\usepackage{amsmath,amsthm,verbatim,amssymb,amsfonts,amscd,mathrsfs}
\usepackage[inline]{enumitem}
\usepackage{color}
\usepackage{wrapfig}
\usepackage{booktabs}
\usepackage{makecell}
\usepackage{multirow}
\ifcomments\newcommand{\comments}[1]{#1}\else\newcommand{\comments}[1]{}\fi
\definecolor{clrgp}{rgb}{.9,0,.9}
\definecolor{blue}{rgb}{0,0,0.5}

\newtheorem{theorem}{Theorem}[section]
\newaliascnt{corollary}{theorem}
\newtheorem{corollary}[corollary]{Corollary}
\aliascntresetthe{corollary}
\newaliascnt{lemma}{theorem}
\newtheorem{lemma}[lemma]{Lemma}
\aliascntresetthe{lemma}

\usepackage{xcolor}

\newcommand{\titl}{Product Kernel Interpolation for Scalable Gaussian Processes}
\newcommand{\titlrunning}{Product Kernel Interpolation for Scalable Gaussian Processes}
\newcommand{\authors}{
  Jacob R. Gardner$^1$, Geoff Pleiss$^1$, Ruihan Wu$^{1,2}$, Kilian Q. Weinberger$^1$, Andrew Gordon Wilson$^1$
}
\newcommand{\institutions}{
  $^1$Cornell University, $^2$Tsinghua University
}

\input{macros}

\begin{document}

\runningtitle{\titlrunning}
\twocolumn[
  \aistatstitle{\titl}
  \aistatsauthor{\authors}
  \aistatsaddress{\institutions}
]

\begin{abstract}
  Recent work shows that inference for Gaussian processes can be performed efficiently using iterative methods that rely only on matrix-vector multiplications (MVMs). Structured Kernel Interpolation (SKI) exploits these techniques by deriving approximate kernels with very fast MVMs. Unfortunately, such strategies suffer badly from the curse of dimensionality.
  We develop a new technique for
 MVM based learning that exploits product kernel structure.  We demonstrate that this technique is broadly applicable, resulting in \emph{linear} rather than exponential runtime with dimension for SKI, as well as state-of-the-art asymptotic complexity for multi-task GPs.
\end{abstract}

\input{sections/introduction}

\input{sections/background}

\input{sections/method}


\input{sections/results}

\input{sections/high_dim}

\input{sections/multi_task}

\input{sections/related_work}

\input{sections/discussion}

\subsubsection*{Acknowledgments}
JRG, GP, and KQW are supported in part by grants from the National Science Foundation (III-1525919, IIS-1550179, IIS-1618134, S\&AS 1724282,
and CCF-1740822), the Office of Naval Research DOD (N00014-17-1-2175), and the Bill and Melinda Gates Foundation.
AGW is supported by NSF IIS-1563887.


\bibliographystyle{apalike}
\bibliography{citations}

%
%


\makeatletter
  \setcounter{table}{0}
  \renewcommand{\thetable}{S\arabic{table}}%
  \setcounter{figure}{0}
  \renewcommand{\thefigure}{S\arabic{figure}}%
  \setcounter{section}{0}
  \renewcommand{\thesection}{S\arabic{section}}
  \renewcommand{\thesubsection}{\thesection.\arabic{subsection}}
  \setcounter{equation}{0}
  \renewcommand\theequation{S\arabic{equation}}
  \renewcommand{\bibnumfmt}[1]{[S#1]}
\makeatother
\newcommand{\suptitl}{Supplementary Materials for:\\ \titl}
\newcommand{\suptitlrunning}{Supplementary Materials for:\\ \titlrunning}

\twocolumn[
  \aistatstitle{\suptitl}
  \aistatsauthor{\authors}
  \aistatsaddress{\institutions}
]
\runningtitle{\suptitlrunning}
\section{Proof of \autoref{lemma:low_rank_mvm}}
Letting $\bq^{(1)}_{i}$ denote the $i^{th}$ row of $Q^{(1)}$ and $\bq^{(2)}_{i}$ denote the $i^{th}$ row of $Q^{(2)}$, we can express the $i^{th}$ entry $\K\bv$, $[\K\bv]_{i}$ as:
\begin{equation} \nonumber
 [\K\bv]_{i} = \bq^{(1)}_{i}\T^{(1)}\Q^{(1)\top} \: D_{\bv} \: \Q^{(2)}\T^{(2)}\bq^{(2)\top}_{i}
\end{equation}
To evaluate this for all $i$, we first once compute the $k \times k$ matrix:
\begin{equation} \nonumber
  M^{(1,2)} = \T^{(1)}\Q^{(1)\top} \: D_{\bv} \: \Q^{(2)}\T^{(2)}.
\end{equation}
This can be done in $O(nk^{2})$ time. $\T^{(1)}\Q^{(1)\top}$ and $\Q^{(2)}\T^{(2)}$ can each be computed in $O(nk^{2})$ time, as the $\Q$ matrices are $n \times k$ and the $\T$ matrices are $n \times k$. Multiplying one of the results by $D_{\bv}$ takes $\bigo{nk}$ time as it is diagonal. Finally, multiplying the two resulting $n \times k$ matrices together takes $\bigo{nk^{2}}$ time.

After computing $M^{(1,2)}$, we can compute each element of the matrix-vector multiply as:
\begin{equation} \nonumber
 [\K\bv]_{i} = \bq^{(1)}_{i}M^{(1,2)}\bq^{(2)\top}_{i}.
\end{equation}
Because $M^{(1,2)}$ is $k \times k$, each of these takes $\bigo{k}$ time to compute. Since there are $n$ entries to evaluate in the MVM $\K\bv$ in total, the total time requirement after computing $M^{(1,2)}$ is $\bigo{kn}$ time. Thus, given low rank structure, we can compute $\K\bv$ in $\bigo{k^{2}n}$ time total.

\section{Proof of \autoref{theorem:main_running_time}}
Given the Lanczos decompositions of $\tilde{K}^{(1)} = \K^{(1)}_{\X\X} \circ \cdots \circ \K^{(a)}_{\X\X}$ and $\tilde{K}^{(2)} = \K^{(a+1)}_{\X\X} \circ \cdots \circ \K^{(d)}_{\X\X}$, we can compute matrix-vector multiplies with $\tilde{K}^{(1)} \circ \tilde{K}^{(2)}$ in $\bigo{k^{2}n}$ time each. This lets us compute the Lanczos decomposition of $\tilde{K}^{(1)} \circ \tilde{K}^{(2)}$ in $\bigo{k^{3}n}$ time.

For clarity, suppose first that $d=3$, i.e., $\K = \K^{(1)}_{\X\X} \circ \K^{(2)}_{\X\X} \circ \K^{(3)}_{\X\X}$. We first Lanczos decompose $\K^{(1)}_{\X\X}$, $\K^{(2)}_{\X\X}$ and $\K^{(3)}_{\X\X}$. Assuming for simplicty that MVMs with each matrix take the same amount of time, This takes $\bigo{k\mvm{\K^{(i)}_{\X\X}}}$ time total. We then use these Lanczos decompositions to compute matrix-vector multiples with $\tilde{K}^{(1)}_{\X\X}$ in $\bigo{k^{2}n}$time each. This allows us to Lanczos decompose it in $\bigo{k^{3}n}$ time total. We can then compute matrix-vector multiplications $\K\bv$ in $\bigo{k^{2}n}$ time.

In the most general setting where $\K=\K^{(1)}_{\X\X} \circ \cdots \circ \K^{(d)}_{\X\X}$, we first Lanczos decompose the $d$ component matrices in $\bigo{dk\mvm{\K^{(i)}}}$ and then perform $\bigo{\log d}$ merges as described above, each of which takes $\bigo{k^{3}n}$ time. After computing all necessary Lanczos decompositions, matrix-vector multiplications with $\K$ can be performed in $\bigo{k^{2}n}$ time.

As a result, a single matrix-vector multiply with $\K$ takes $\bigo{dk\mvm{\K^{(i)}} + k^{3}n\log d + k^{2}n}$ time. With the Lanczos decompositions precomputed, multiple MVMs in a row can be performed significantly faster. For example, running $p$ iterations of conjugate gradients with $\K$ takes $\bigo{dk\mvm{\K^{(i)}} + k^{3}n\log d + pk^{2}n}$ time.
\end{document}

%% file: macros.tex
\newcommand{\reals}{\ensuremath{\mathbb{R}}}

\newcommand{\bigo}[1]{\mathcal O ( #1 )}
\newcommand{\mvm}[1]{\mu ( #1 )}
\newcommand{\set}[1]{\left\{ #1 \right\}}

\newif\ifboldmatrix
\boldmatrixfalse
\ifboldmatrix\newcommand{\boldmatrix}[1]{\mathbf{#1}}\else\newcommand{\boldmatrix}[1]{#1}\fi

\newcommand{\x}{\ensuremath{\mathbf{x}}}
\newcommand{\bb}{\ensuremath{\mathbf{b}}}
\newcommand{\bv}{\ensuremath{\mathbf{v}}}
\newcommand{\bq}{\ensuremath{\mathbf{q}}}
\newcommand{\bw}{\ensuremath{\mathbf{w}}}

\newcommand{\X}{\ensuremath{\boldmatrix{X}}}

\newcommand{\W}{\ensuremath{\boldmatrix{W}}}
\newcommand{\B}{\ensuremath{\boldmatrix{B}}}
\newcommand{\A}{\ensuremath{\boldmatrix{A}}}

\newcommand{\Q}{\ensuremath{\boldmatrix{Q}}}
\newcommand{\T}{\ensuremath{\boldmatrix{T}}}
\newcommand{\U}{\ensuremath{\boldmatrix{U}}}

\newcommand{\K}{\ensuremath{\boldmatrix{K}}}
\newcommand{\y}{\ensuremath{\mathbf{y}}}
\newcommand{\bz}{\ensuremath{\mathbf{z}}}
\newcommand{\fn}{\ensuremath{\mathbf{f}}}

\newcommand{\dset}{\ensuremath{\mathcal D}}

\newcommand{\numpeople}{{\ensuremath{s}}}

%% file: sections/introduction.tex
\section{INTRODUCTION}

Gaussian processes (GPs) provide a powerful approach to regression and extrapolation, with applications as varied as
time series analysis \citep{wilson2013gaussian,DuvLloGroetal13}, blackbox optimization \citep{jones1998efficient,snoek2012practical}, and personalized medicine and counterfactual prediction \citep{durichen2015multitask, schulam2015framework, herlands2016scalable,gardner2015psychophysical}. Historically, one of the key limitations of Gaussian process regression has been the computational intractability of inference when dealing with more than a few thousand data points. This complexity stems from the need to solve linear systems and compute log determinants involving an $n \times n$ symmetric positive definite \emph{covariance matrix} $\K$. This task is commonly performed by computing the Cholesky decomposition of $K$ \citep{rasmussen2006gaussian}, incurring $\bigo{n^{3}}$ complexity. To reduce this complexity, \emph{inducing point methods} make use of a small set of $m < n$ points to form a rank $m$ approximation of $\K$ \citep{quinonero2005unifying, snelson2006sparse,hensman2013gaussian,titsias2009variational}. Using the matrix inversion and determinant lemmas, inference can be performed in $\bigo{nm^{2}}$ time \citep{snelson2006sparse}.

Recently, however, an alternative class of inference techniques for Gaussian processes have emerged based on iterative numerical linear algebra techniques \citep{wilson2015kernel,dong2017scalable}. Rather than explicitly decomposing the full covariance matrix, these methods leverage Krylov subspace methods \citep{golub2012matrix} to perform linear solves and log determinants using only matrix-vector multiples (MVMs) with the covariance matrix.
Letting $\mvm{\K}$ denote the time complexity of computing $\K\bv$ given a vector $\bv$, these methods provide excellent approximations to linear solves and log determinants in $\bigo{r\mvm{\K}}$ time, where $r$ is typically some small constant \cite{golub2012matrix}.\footnote{
  In practice, $r$ depends on the conditioning of $K$, but is independent of $n$.}
This approach has led to scalable GP methods that differ radically from previous approaches -- the goal shifts from computing efficient Cholesky decompositions to computing efficient MVMs. Structured kernel interpolation (SKI) \citep{wilson2015kernel} is a recently proposed inducing point method that, given a regular grid of $m$ inducing points, allows for MVMs to be performed in an impressive $\bigo{n + m \log m}$ time.    

These MVM approaches have two fundamental drawbacks. First, \citet{wilson2015kernel} use Kronecker factorizations for SKI to take advantage of fast MVMs, constraining the number of inducing points $m$ to grow exponentially with the dimensionality of the inputs, limiting the applicability of SKI to problems with fewer than about $5$ input dimensions.
Second, the computational benefits of iterative MVM inference methods come at the cost of reduced modularity. If all we know about a kernel is that it decomposes as $\K = \K_{1} \circ \K_{2}$, it is not obvious how to efficiently perform MVMs with $\K$, even if we have access to fast MVMs with both $\K_{1}$ and $\K_{2}$.
In order for MVM inference to be truly modular, we should be able to perform inference equipped with nothing but the ability to perform MVMs with $K$.   One of the primary advantages of GPs is the ability to construct very expressive kernels by composing simpler ones \citep{rasmussen2006gaussian, gonen2011multiple, durrande2011additive, DuvLloGroetal13, wilson2014covariance}. One of the most common kernel compositions is the element-wise product of kernels. This composition can encode different functional properties for each input dimension \citep[e.g.,][]{rasmussen2006gaussian, gonen2011multiple, DuvLloGroetal13, wilson2014covariance}, or express correlations between outputs in multi-task settings \citep{mackay98, bonilla2008multi, alvarez2011computationally}.  Moreover, the RBF and ARD kernels -- arguably the most popular kernels in use -- decompose into product kernels.

In this paper, we propose a single solution which addresses
both of these limitations of iterative methods -- improving modularity while simultaneously alleviating the curse of dimensionality. In particular:





\begin{enumerate}[wide, labelwidth=!, labelindent=0pt]
    \item We demonstrate that MVMs with product kernels can be approximated efficiently by computing the Lanczos decomposition of each component kernel. If MVMs with a kernel $\K$ can be performed in $\bigo{\mvm{\K}}$ time, then MVMs with the element-wise product of $d$ kernels can be approximated in $\bigo{dr\mvm{\K} + r^{3}n\log d}$ time, where $r$ is typically a very small constant.
    \item Our fast product-kernel MVM algorithm, entitled \emph{SKIP}, enables the use of structured kernel interpolation with product kernels without resorting to the exponential complexity of Kronecker products.  SKIP can be applied even when the product kernels use different interpolation grids, and enables GP inference and learning in $\bigo{dn + d m \log m}$ for products of $d$ kernels.
    \item We apply SKIP to high-dimensional regression problems by expressing $d$-dimensional kernels as the product of $d$ one-dimensional kernels. This formulation affords an \emph{exponential improvement} over the standard SKI complexity of $\bigo{n + dm^{d} \log m}$, and achieving state of the art performance over popular inducing point methods \citep{hensman2013gaussian, titsias2009variational}.
    \item We demonstrate that SKIP can reduce the complexity of multi-task GPs (MTGPs) to $\bigo{n + m \log m + s}$ for a problem with $s$ tasks. We exploit this fast inference, developing a model that discovers cluster of tasks using Gibbs sampling.
    \item We make our GPU implementations available as easy to use code as part of a new package for Gaussian processes, GPyTorch, available at \url{https://github.com/cornellius-gp/gpytorch}.
\end{enumerate}

%% file: sections/background.tex
\section{BACKGROUND}
In this section, we provide a brief review of Gaussian process regression and an overview of iterative inference techniques for Gaussian processes based on matrix-vector multiplies.

\subsection{Gaussian Processes}
A Gaussian process generalizes multivariate normal distributions to distributions over functions that are specified by a prior \emph{mean function} and a prior \emph{covariance function} $f(\x)\sim\mathcal{GP} \left(\mu(\x),k(\x,\x')\right)$. By definition, the function values of a GP at any finite set of inputs $[\x_{1},...,\x_{n}]$ are jointly Gaussian distributed:
\begin{equation*}
    \fn = [f(\x_{1}),...,f(\x_{n})] \sim\mathcal{N} \left( \mu_{\X}, \K_{\X\X} \right)
\end{equation*}
where $\mu_{\X} = [\mu(\x_{1}),...,\mu(\x_{n})]^{\top}$ and $\K_{\X\X}=[k(\x_{i},\x_{j})]_{i,j=1}^{n}$. Generally, $K_{AB}$ denotes a matrix of cross-covariances between the sets $A$ and $B$.


Under a Gaussian noise observation model, $p(y(\x) \mid f(\x)) \sim \mathcal{N}(y(\x); f(\x),\sigma^2)$,
the predictive distribution at $\x^{*}$ given data $\mathcal{D} = \{(\x_i, y_i)\}_{i=1}^{n}$ is
\begin{align}
    p(f(\x^{*})\mid\dset) &\sim \mathcal{GP} \left(\mu_{f\mid\dset}(\x^{*}), k_{f\mid\dset}(\x^{*}, \x^{*'})\right), \notag \\
\mu_{f\mid\dset}(\x) &= \mu(\x^{*}) + \K_{\x^{*}\X}\hat{K}_{\X\X}^{-1} \y \label{eq:pred_mean}, \\
  k_{f\mid\dset}(\x^*, \x^*) &= \K_{\x^{*}\x^{*}} - \K_{\x^{*}\X}\hat{K}_{\X\X}^{-1}\K_{\x^{*}\X}^{\top}, \label{eq:pred_covar}
\end{align}
where $\hat{K}_{\X\X} = K_{\X\X} + \sigma^2 I$ and $\y = (y(\x_1),\dots,y(\x_n))^{\top}$.
All kernel matrices implicitly depend on hyperparameters $\theta$.  The log \emph{marginal likelihood} of the data, conditioned only on these hyperparameters, is given by
\begin{equation}
  \label{eq:marginalloglik}
    \log p(\y\mid\theta) = -\frac{1}{2}\y^{\top}\hat{K}_{\X\X}^{-1}\y - \frac{1}{2} \log \vert \hat{K}_{\X\X} \vert + \text{c} \,,
\end{equation}
which provides a utility function for kernel learning.

\subsection{Inference with matrix-vector multiplies}
\label{sec:mvm}
In order to compute the predictive mean in \eqref{eq:pred_mean}, the predictive covariance in \eqref{eq:pred_covar}, and the marginal log likelihood in \eqref{eq:marginalloglik}, we need to perform
linear solves (i.e. $[\K_{\X\X} + \sigma^{2} I]^{-1}\bv$) and log determinants (i.e. $\log | \K_{\X\X} + \sigma^{2} I |$).
Traditionally, these operations are achieved using the Cholesky decomposition of $\K_{\X\X}$ \citep{rasmussen2006gaussian}.
Computing this decomposition requires $\bigo{n^3}$ operations and storing the result requires $\bigo{n^2}$ space.
Given the Cholesky decomposition, linear solves can be computed in $\bigo{n^{2}}$ time and log determinants in $\bigo{n}$ time.

There exist alternative approaches \citep[e.g.][]{wilson2015kernel} that require only matrix-vector multiplies (MVMs) with $[\K_{\X\X} + \sigma^{2} I]$.
To compute linear solves, we use the method of \emph{conjugate gradients} (CG). This technique exploits
that the solution to $\A\x = \bb$ is the unique minimizer of the quadratic function
  $\frac{1}{2} \x^{\top}\A\x - \x^{\top}\bb$,
%
which can be found by iterating a simple three term recurrence.
Each iteration requires a single MVM with the matrix $\A$ \citep{shewchuk1994introduction}.
Letting $\mvm{\A}$ denote the time complexity of an MVM with $\A$, $p$ iterations of CG requires $O(p\mvm{A})$ time. If $\A$ is $n \times n$, then CG is exact when $p=n$. However, the linear solve can often be approximated by $p<n$ iterations, since the magnitude of the residual $\mathbf{r} = \A\x - \bb$ often decays exponentially.
In practice the value of $k$ required for convergence to high precision is a small constant that depends on the conditioning of $\A$ rather than $n$ \citep{golub2012matrix}.
A similar technique known as \emph{stochastic Lanczos quadrature} exists for approximating log determinants in $O(p\mvm{A})$ time \citep{dong2017scalable,ubaru2017fast}.
In short, inference and learning for GP regression can be done in $\bigo{p\mvm{\K_{\X\X}}}$ time using these iterative approaches.

Critically, if the kernel matrices admit fast MVMs -- either through the structure of the data \citep{saatcci2012scalable,cunningham2008fast} or the structure of a general purpose kernel approximation \citep{wilson2015kernel} -- this iterative approach offers massive scalability gains over conventional Cholesky-based methods.

\subsection{Structured kernel interpolation}
\label{sec: ski}

Structured kernel interpolation (SKI) \citep{wilson2015kernel} replaces a user-specified kernel $k(\x,\x')$ with an approximate kernel that affords very fast matrix-vector multiplies.
Assume we are given a set of $m$ \emph{inducing points} $\U$ that we will use to approximate kernel values.
Instead of computing kernel values between data points directly, SKI computes kernel values between inducing points and \emph{interpolates} these kernel values to approximate the true data kernel values. This leads to the approximate SKI kernel:
\begin{equation}\label{eq:interp_single}
    k(\x,\bz) \approx \bw_{\x}\K_{\U\U}\bw_{\bz}^{\top},
\end{equation}
where $\bw_{\x}$ is a sparse vector that contains interpolation weights. For example, when using local cubic interpolation \citep{keys1981cubic}, $\bw_{x}$ contains four nonzero elements. Applying this approximation for all data points in the training set, we see that:
\begin{equation}
  \K_{\X\X} \approx \W_{\X}\K_{\U\U}\W_{\X}^{\top}
\end{equation}
With arbitrary inducing points $\U$, matrix-vector multiplies with $[\W\K_{\U\U}\W^{\top}]\bv$ require $\bigo{n+m^{2}}$ time. In one dimension, we can reduce this running time by instead choosing $\U$ to be a regular grid, which results in $\K_{\U\U}$ being \emph{Toeplitz}. In higher dimensions, a multi-dimensional grid results in $\K_{\U\U}$ being the Kronecker product of Toeplitz matrices. This decompositions enables matrix-vector multiplies in at most $\bigo{n + m \log m}$ time, and $\bigo{n + m}$ storage.  However, a Kronecker decomposition of $\K_{\U\U}$ leads to an exponential time complexity in $d$, the dimensionality of the inputs $\mathbf{x}$ \citep{wilson2015kernel}.




%% file: sections/method.tex
\section{MVMs WITH PRODUCT KERNELS}
\label{sec: skip}
\begin{figure}[t!]
   \centering
   \includegraphics[width=\linewidth]{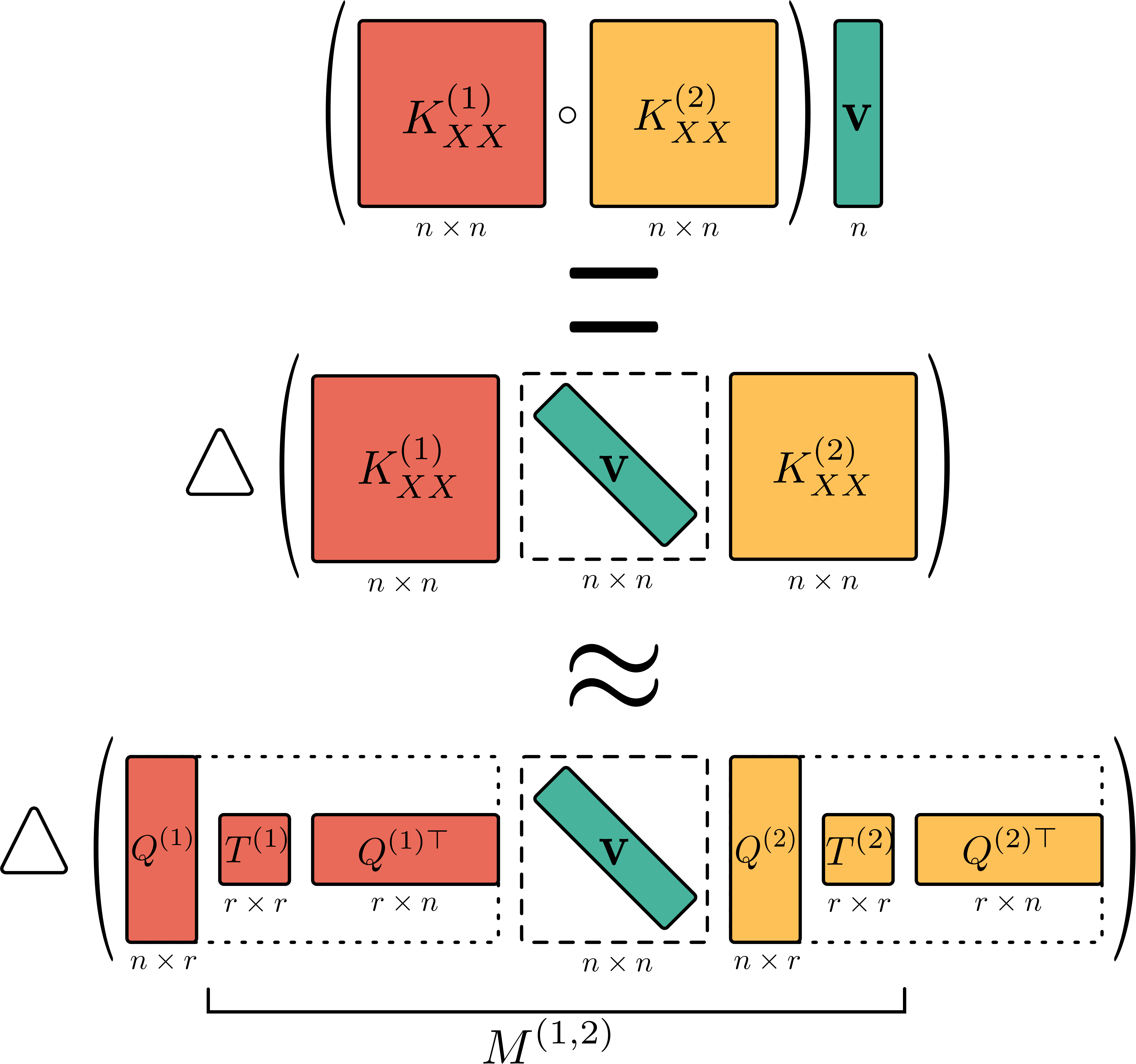}
   \caption{
     Computing fast matrix-vector multiplies (MVMs) with the product kernel $K^{(1)}_{\X\X} \circ K^{(2)}_{\X\X}.$
     {\bf 1:} Rewrite the element-wise product as the diagonal $\Delta(\cdot)$ of a product of matrices.
     {\bf 2:} Compute the rank-$r$ Lanczos decomposition of $K^{(1)}_{\X\X}$ and $K^{(2)}_{\X\X}$.
   }
   \label{fig:product_kernel}
   \raggedbottom
\end{figure}

In this section we derive an approach to exploit product kernel structure for fast MVMs, towards
alleviating the curse of dimensionality in SKI.  Suppose a kernel separates as a product as follows:
%
%
\begin{equation}
    \label{eq:prod_kernel}
    k(\x, \x') = \prod_{i=1}^{d}k^{(i)}(\x, \x').
\end{equation}
%
Given a training data set $\mathbf \X = [ \x_1, \ldots \x_n ]$, the kernel matrix $\K$ resulting from the product of kernels in \eqref{eq:prod_kernel} can be expressed as $\K=\K^{(1)}_{\X\X} \circ \cdots \circ \K^{(d)}_{\X\X}$, where $\circ$ represents element-wise multiplication. In other words:
\begin{equation}
  \left[K^{(1)}_{\X\X} \circ K^{(2)}_{\X\X}\right]_{ij} = \left[K^{(1)}_{\X\X}\right]_{ij}\left[K^{(2)}_{\X\X}\right]_{ij}.
\end{equation}
%
%
%
The key limitation we must deal with is that, unlike a sum of matrices, vector multiplication does not distribute over the elementwise product:
\begin{equation}
  \left( K^{(1)} \circ K^{(2)} \right) \bv \ne \left( K^{(1)} \bv \right) \circ \left( K^{(2)} \bv \right).
\end{equation}
%


%

We will assume we have access to fast MVMs for each component kernel matrix $K^{(i)}$. Without fast MVMs, there is a trivial solution to computing the elementwise matrix vector product: explicitly compute the kernel matrix $\K$ in $\bigo{dn^{2}}$ time and then compute $\K\bv$. We further assume that $K^{(i)}$ admits a low rank approximation, following prior work on inducing point methods following prior work on inducing point methods \citep{snelson2006sparse,titsias2009variational,wilson2015kernel,hensman2013gaussian}.

\paragraph{A naive algorithm for a two-kernel product.}
We initially assume for simplicity that there are only $d=2$ components kernels in the product.
We will then show how to extend the two kernel case to arbitrarily sized product kernels. We seek to perform matrix vector multiplies:
\begin{equation}
  \label{eq:mvm_2_terms}
  (K^{(1)}_{\X\X} \circ K^{(2)}_{\X\X})\bv
\end{equation}
Eq.~\eqref{eq:mvm_2_terms} may be expressed in terms of matrix-matrix multiplication using the following identity:
\begin{equation}
    \label{eq:two_mat_hpvm}
    \K\bv = (K^{(1)}_{\X\X} \circ K^{(2)}_{\X\X})\bv = \Delta(K^{(1)}_{\X\X} \: D_{\bv} \: K^{(2)\top}_{\X\X}),
\end{equation}
where $D_{\bv}$ is a diagonal matrix whose elements are $\bv$ (\autoref{fig:product_kernel}), and $\Delta(M)$ denotes the diagonal of $M$. Because $D_{\bv}$ is an $n \times n$ matrix, computing the entries of $K \bv$ naively requires $n$ matrix-vector multiplies with $K^{(1)}_{\X\X}$ and $K^{(2)}_{\X\X}$. The time complexity to compute \eqref{eq:two_mat_hpvm} is therefore $\bigo{n\mvm{K^{(1)}_{\X\X}} + n\mvm{K^{(2)}_{\X\X}}}$. This reformulation does not naively offer any time savings.
\paragraph{Exploiting low-rank structure.}
Suppose however that we have access to rank-$r$ approximations of $\K^{(1)}_{\X\X}$ and $\K^{(2)}_{\X\X}$:
%
\begin{equation}\nonumber
  K^{(1)}_{\X\X} \approx \Q^{(1)}\T^{(1)}\Q^{(1)\top},
  \:\:\:\:\:
  K^{(2)}_{\X\X} \approx \Q^{(2)}\T^{(2)}\Q^{(2)\top},
\end{equation}
where $\Q^{(1)}$, $\Q^{(2)}$ are $n \times r$ and $\T^{(1)}$, $\T^{(2)}$ are $r \times r$ (\autoref{fig:product_kernel}).
This rank decomposition makes the MVM significantly cheaper to compute.
Plugging these decompositions in to \eqref{eq:two_mat_hpvm}, we derive:
\begin{equation}
  \label{eq:two_mat_hpvm_low_rank}
 \K\bv = \Delta\left(\Q^{(1)}\T^{(1)}\Q^{(1)\top} \: D_{\bv} \: \Q^{(2)}\T^{(2)}\Q^{(2)\top}\right).
\end{equation}
We prove the following key lemma in the supplementary materials about \eqref{eq:two_mat_hpvm_low_rank}:
\begin{lemma}
\label{lemma:low_rank_mvm}
Suppose that $\K^{(1)}_{\X\X}=\Q^{(1)}\T^{(1)}\Q^{(1)\top}$ and $\K^{(2)}_{\X\X}=\Q^{(1)}\T^{(1)}\Q^{(1)\top}$, where $\Q^{(1)}$ and $\Q^{(2)}$ are $n \times r$ matrices and $\T^{(1)}$ and $T^{(2)}$ are $r \times r$. Then $(\K^{(1)}_{\X\X} \circ \K^{(2)}_{\X\X})\bv$ can be computed with \eqref{eq:two_mat_hpvm_low_rank} in $\bigo{r^{2}n}$ time.
\end{lemma}
Therefore, if we can efficiently compute low-rank decompositions of $\K^{(1)}$ and $\K^{(2)}$, then we immediately apply \autoref{lemma:low_rank_mvm} to perform fast MVMs.

\paragraph{Computing low-rank structure.}
With \autoref{lemma:low_rank_mvm}, we have reduced the problem of computing MVMs with $\K$ to that of constructing low-rank decompositions for $K^{(1)}_{\X\X}$ and $K^{(2)}_{\X\X}$.  Since we are assuming we
can take fast MVMs with these kernel matrices, we now turn to the \emph{Lanczos decomposition} \citep{lanczos1950iteration,paige1972computational}.
The Lanczos decomposition is an iterative algorithm that takes a symmetric matrix $A$ and probe vector $b$ and returns $Q$ and $T$ such that $A \approx QTQ^{\top}$, with $\Q$ orthogonal.

This decomposition is exact after $n$ iterations. However, if we only compute $r < n$ columns of $Q$, then $Q_{r}T_{r}Q_{r}^{\top}$ is an effective low-rank approximation of $A$ \citep{nickisch2009bayesian,simon2000low}. Unlike standard low rank approximations (such as the singular value decomposition), the algorithm for computing the Lanczos decomposition $\K^{(i)}_{\X\X} = Q^{(i)}T^{(i)}Q^{(i)\top}$ requires only $r$ MVMs, leading to the following lemma:
\begin{lemma}
\label{lemma:lanczos_time}
Suppose that MVMs with $\K^{(i)}_{\X\X}$ can be computed in $\bigo{\mvm{\K^{(i)}_{\X\X}}}$ time. Then the rank-$r$ Lanczos decomposition $\K^{(i)}_{\X\X} \approx \Q_r^{(i)}\T_r^{(i)}\Q_r^{(i)\top}$ can be computed in $\bigo{r \mvm{\K^{(i)}_{\X\X}}}$ time.
\end{lemma}
The above discussion motivates the following algorithm for computing $(K^{(1)}_{\X\X} \cdot K^{(2)}_{\X\X})\bv$, which is summarized by \autoref{fig:product_kernel}: First, compute the rank-$r$ Lanczos decomposition of each matrix;
then, apply \eqref{eq:two_mat_hpvm_low_rank}. Lemmas \ref{lemma:lanczos_time} and \ref{lemma:low_rank_mvm} together imply that this takes $\bigo{r\mvm{\K^{(1)}_{\X\X}} + r\mvm{\K^{(2)}_{\X\X}} + r^{2}n}$ time.
\begin{figure*}[t!]
  \centering
  \includegraphics[width=0.8\columnwidth]{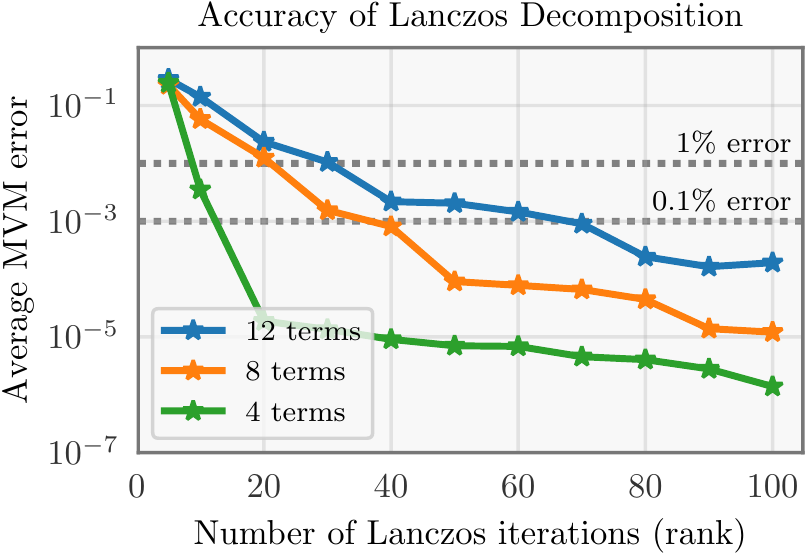}
  \hspace{3ex}
  \includegraphics[width=0.8\columnwidth]{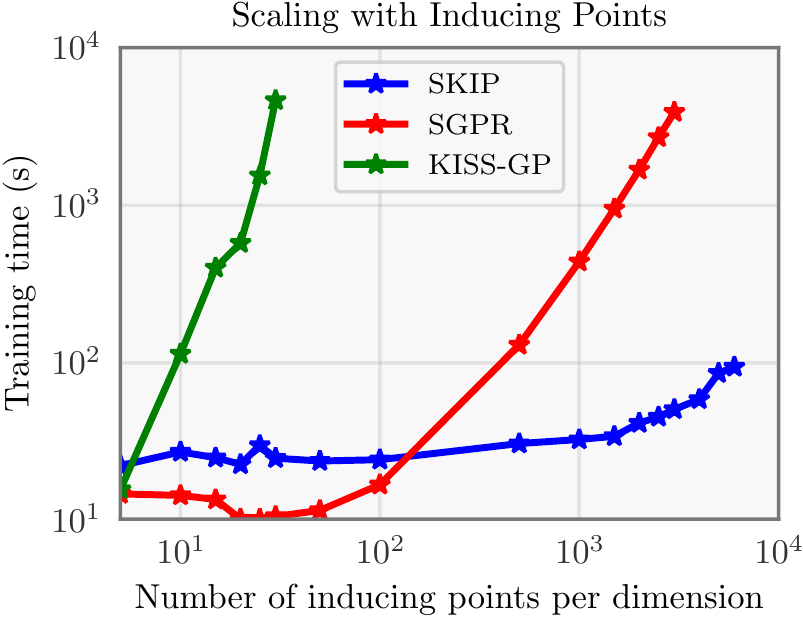}
  \caption{{\bf Left:} Relative error of MVMs computed using SKIP compared to the exact value $\K\bv$. {\bf Right:} Training time as a function of the number of inducing points \emph{per dimension}.   KISS-GP (SKI with Kronecker factorization) scales well with the \emph{total} number of inducing points, but badly with number of inducing points \emph{per dimension}, because the required total number of inducing points scales exponentially with the number of dimensions. \label{fig:mvm_evaluation}}
  \vspace{-2ex}
\end{figure*}

\paragraph{Extending to product kernels with three components.}
Now consider a kernel that decomposes as the product of three components, $k(\x, \x') = k^{(1)}(\x, \x') k^{(2)}(\x, \x')k^{(3)}(\x, \x')$. An MVM with this kernel is given by
$\K\bv = (\K^{(1)}_{\X\X} \circ \K^{(2)}_{\X\X} \circ \K^{(3)}_{\X\X})\bv.$
Define $\tilde{K}^{(1)}_{\X\X} = \K^{(1)}_{\X\X} \circ \K^{(2)}_{\X\X}$ and $\tilde{K}^{(2)}_{\X\X} = \K^{(3)}_{\X\X}$. Then
\begin{equation}
  \label{eq:three_mat_case}
  \K\bv = (\tilde{K}^{(1)}_{\X\X} \circ \tilde{K}^{(2)}_{\X\X})\bv \,,
\end{equation}
reducing the three component problem back to two components. To compute the Lanczos decomposition of $\tilde{K}^{(1)}_{\X\X}$, we use the method described above for computing MVMs with $\K^{(1)}_{\X\X} \circ \K^{(2)}_{\X\X}$.
\paragraph{Extending to product kernels with many components.}
The approach for the three component setting leads naturally to a divide and conquer strategy. Given a kernel matrix $\K = \K^{(1)}_{\X\X} \circ \cdots \circ \K^{(d)}_{\X\X}$ we define
\begin{align}
\tilde{K}^{(1)}_{\X\X}&=\K^{(1)}_{\X\X} \circ \cdots \circ \K^{(\frac{d}{2})}_{\X\X} \\
\tilde{K}^{(2)}_{\X\X}&=\K^{(\frac{d}{2}+1)}_{\X\X} \circ \cdots \circ \K^{(d)}_{\X\X},
\end{align}
which lets us rewrite $\K=\tilde{K}^{(1)}_{\X\X} \circ \tilde{K}^{(2)}_{\X\X}$. By applying this splitting recursively, we can compute matrix-vector multiplies with $\K$, leading to the following running time complexity:
\begin{theorem}
  \label{theorem:main_running_time}
  Suppose that $\K=\K^{(1)}_{\X\X} \circ \cdots \circ \K^{(d)}_{\X\X}$, and that computing a matrix-vector multiply with any $\K^{(i)}_{\X\X}$ requires $\bigo{\mvm{\K^{(1)}_{\X\X}}}$ operations. Computing an MVM with $\K$ requires $\bigo{dr\mvm{\K^{(i)}} + r^{3}n\log d + r^{2}n}$ time, where $r$ is the rank of the Lanczos decomposition used.
\end{theorem}
\paragraph{Sequential MVMs.}
If we are computing many MVMs with the same matrix, then we can further reduce this complexity by caching the Lanczos decomposition.
The terms $\bigo{dr\mvm{\K^{(i)}_{\X\X}} + r^{3}n \log d}$ represent the time to construct the Lanczos decomposition.
However, note that this decomposition is not dependent on the vector that we wish to multiply with. Therefore, if we save the decomposition for future computation, we have the following corollary:
\begin{corollary}
  Any subsequent MVMs with $\K$ require $\bigo{r^{2}n}$ time.
\end{corollary}
 If matrix-vector multiplications with $\K^{(i)}_{\X\X}$ can be performed with significantly fewer than $n^{2}$ operations, this results in a significant complexity improvement over explicitly computing the full kernel matrix $\K$.

\subsection{Structured kernel interpolation for products (SKIP)}
So far we have assumed access to fast MVMs with each constituent kernel matrix of an elementwise (Hadamard) product:
$\K=\K^{(1)}_{\X\X} \circ \cdots \circ \K^{(d)}_{\X\X}$.  To achieve this, we apply the SKI approximation (Section~\ref{sec: ski}) to each component:
\begin{equation}
\K^{(i)}_{\X\X}=\W^{(i)}\K_{\U\U}\W^{(i)\top}.
\end{equation}
\emph{}When using SKI approximations, the running time of our product kernel inference technique with $p$ iterations of CG becomes $\bigo{dr(n + m \log m) + r^{3}n\log d + pr^{2}n}$. The running time of SKIP is compared to that of other inference techniques in \autoref{tab:time}.

%% file: sections/results.tex
\vspace{-2ex}
\section{MVM ACCURACY AND SCALABILITY}
\vspace{-2ex}
We first evaluate the accuracy of our proposed approach with product kernels in a controlled synthetic setting. We draw $2500$ data points in $d$ dimensions from $\mathcal{N}(0, I)$ and compute an RBF kernel matrix with lengthscale $1$ over these data points.
We evaluate the relative error of SKIP compared exact MVMs as a function of $r$ -- the number of Lanczos iterators. We perform this test for for 4, 8, and 12 dimensional data, resulting in a product kernel with 4, 8, and 12 components respectively.
The results, averaged over 100 trials, are shown in \autoref{fig:mvm_evaluation} (left). Even in the 12 dimensional setting, an extremely small value of $r$ is sufficient to get very accurate MVMs: less than 1\% error is achieved when $k=30$. For a discussion of increasing error with dimensionality, see \autoref{sec:discussion}. In future experiments, we set the maximum number of Lanczos iterations to $100$, but note that the convergence criteria is typically met far sooner. In the right side of \autoref{fig:mvm_evaluation}, we demonstrate the improved scaling of our method with the number of inducing points per dimension over KISS-GP. To do this, we use the $d=4$ dimensional Power dataset from the UCI repository, and plot inference step time as a function of $m$. While our method clearly scales better with $m$ than both KISS-GP and SGPR, we also note that because SKIP only applies the inducing point approximation to one-dimensional kernels, we anticipate ultimately needing significantly fewer inducing points than either SGPR or KISS-GP which need to cover the full $d$ dimensional space with inducing points.

%% file: sections/high_dim.tex
\begin{table*}[t!]
  \caption{Comparison of SKIP and other methods on higher dimensional datasets.  In this table, $m$ is the \emph{total} number of inducing points, rather than number of inducing points per dimension.
  (*We use $m=100$ for SKIP on all datasets except precipitation, where we use $m=120 \text K$.) \label{tab:highd_results}}
  \centering
  \vspace{3pt}
  \resizebox{\textwidth}{!}{
    \input results/highd_results
  }
  \vspace{-2ex}
\end{table*}

\section{APPLICATION 1: AN EXPONENTIAL IMPROVEMENT TO SKI}
\begin{table}
  \caption{Asymptotic complexities of a single calculation of \autoref{eq:marginalloglik} with $n$ data points, $m$ inducing points, $r$ Lanczos iterations and $p$ CG iterations. The first two rows correspond to an exact GP with Cholesky and CG. \label{tab:time}}
  \vspace{0.5ex}
  \centering
  \resizebox{\columnwidth}{!}{%
  \input results/time_complexities.tex
  }
  \vspace{-2ex}
\end{table}
\citet{wilson2015kernel} use a Kronecker decomposition of $K_{UU}$ to apply SKI for $d > 1$ dimensions, which requires a fully connected multi-dimensional grid of inducing points $U$.  Thus if we wish to have $m$ distinct inducing point values for each dimension, the grid requires $m^{d}$ inducing points -- i.e. MVMs with the SKI approximate $\K_{\X\X}$ require $\bigo{n + dm^{d}\log m}$ time.  It is therefore computationally infeasible to apply SKI with a Kronecker factorization, referred to in \citet{wilson2015kernel} as KISS-GP, to more than about five dimensions. 
However, using the proposed SKIP method of Section~\ref{sec: skip}, we can reduce the running time complexity of SKI in $d$ dimensions from exponential $\bigo{n + dm^{d}\log m}$ to linear $\bigo{dn + dm\log m}$!
If we express a $d$-dimensional kernel as the product of $d$ one-dimensional kernels, then each component kernel requires only $m$ grid points, rather than $m^{d}$.
For the RBF and ARD kernels, decomposing the kernel in this way yields the same kernel function.
%
%

\paragraph{Datasets.} We evaluate SKIP on six benchmark datasets. The precipitation dataset contains hourly rainfall measurements from hundreds of stations around the country. The remaining datasets are taken from the UCI machine learning dataset repository. KISS-GP (SKI with a Kronecker factorization) is not applicable when $d>5$, and the full GP is not applicable on the four largest datasets.

\paragraph{Methods.} We compare against the popular sparse variational Gaussian processes (SGPR)
\citep{titsias2009variational, hensman2013gaussian} implemented in GPflow \citep{matthews2017gpflow}. We also compare to our GPU implementation of KISS-GP where possible, as well as our GPU implementation of the full GP on the two smallest datasets. All experiments were run on an NVIDIA Titan Xp. We evaluate SGPR using 200, 400 and 800 inducing points. All models use the RBF kernel and a constant prior mean function. We optimize hyperparameters with ADAM using default optimization parameters.

\paragraph{Discussion.} The results of our experiments are shown in \autoref{tab:highd_results}. On the two smallest datasets, the Full GP model outperforms all other methods in terms of speed. This is due to the overhead added by inducing point methods significantly outweighing simple solves with conjugate gradients with such little data. SKIP is able to match the error of the full GP model on elevators, and all methods have comparable error on the Pumadyn dataset.

On the precipitation dataset, inference with standard KISS-GP is still tractable due to the low dimensionality, and KISS-GP is both fast and accurate. Using SKIP results in higher error than KISS-GP, because we were able to use significantly fewer Lanczos iterations for our approximate MVMs than on other datasets due to the space complexity. We discuss the space complexity limitation further in the discussion section. Nevertheless, SKIP still performs better than SGPR. SGPR results with 400 and 800 inducing points are unavailable due to GPU memory constraints. On the remaining datasets, SKIP is able to achieve comparable or better overall error than SGPR, but with a significantly lower runtime.

%% file: results/highd_results.tex
\begin{tabular}{ |c|c||c||c|c|c|c||c| }
  \hline
  {\bf Dataset} &
  {\bf Metric} &
  \thead{{\bf Full GP}} &
  \thead{{\bf SGPR} \\ ($m=200$)} &
  \thead{{\bf SGPR} \\ ($m=400$)} &
  \thead{{\bf SGPR} \\ ($m=800$)} &
  \thead{{\bf KISS-GP} \\ ($m=120 \text K$)} &
  \thead{{\bf SKIP} \\ ($m=100$)\textsuperscript{*}}
  \\
  \hline

    \multirow{2}*{\bf \thead{Pumadyn \\ ($n=8192$, $d=32$)}} &
      \bf Test MAE &
      \bf 0.721    &
      0.766    &
      0.766    &
      0.766     &
      --      &
      0.766
      \\

      & \bf Train Time ($s$) &
      \bf 4  &
      28 &
      67 &
      235 &
      -- &
      65

    \\
    \hdashline

    \multirow{2}*{\bf \thead{Elevators \\ ($n=16599$, $d=18$)}} &
      \bf Test MAE &
      \bf 0.072    &
      0.157    &
      0.157     &
      0.157     &
      --      &
      \bf0.072
      \\

      & \bf Train Time ($s$) &
      \bf 12 &
      46 &
      122 &
      425 &
      -- &
      23

    \\
    \hdashline

  \multirow{2}*{\bf \thead{Precipitation \\ ($n=628474$, $d=3$)}} &
    \bf Test MAE &
    --   &
    14.79 &
    --   &
    --   &
    \bf 9.81 &
    14.08
    \\

    & \bf Train Time ($s$) &
    -- &
    1432 &
    -- &
    -- &
    615 &
    \bf 34.16

  \\
  \hdashline

 \multirow{2}*{\bf \thead{KEGG \\ ($n=48827$, $d=22$)}} &
    \bf Test MAE &
    --      &
    0.101    &
    0.093     &
    0.087     &
    --      &
    \bf0.065
    \\

    & \bf Train Time ($s$) &
    -- &
    116 &
    299 &
    9926 &
    -- &
    \bf 66

  \\
  \hdashline

  \multirow{2}*{\bf \thead{Protein \\ ($n=45730$, $d=9$)}} &
    \bf Test MAE &
    --      &
    7.219    &
    4.97     &
    4.72     &
    --      &
    \bf1.97
    \\

    & \bf Train Time ($s$) &
    -- &
    139 &
    397 &
    1296 &
    -- &
    \bf 35

  \\
  \hdashline

  \multirow{2}*{\bf \thead{Video \\ ($n=68784$, $d=16$)}} &
    \bf Test MAE &
    --   &
    6.836 &
    6.463 &
    6.270 &
    -- &
    \bf 5.621
    \\

    & \bf Train Time ($s$) &
    -- &
    113 &
    334 &
    1125 &
    --  &
    \bf 57

  \\
  \hline
\end{tabular}

%% file: results/time_complexities.tex
\begin{tabular}{ |c|c| }
  \hline
  {\bf Method} & {\bf Complexity of 1 Inference Step} \\
  \hline
  GP (Chol) & $\bigo{n^{3}}$ \\
  GP (MVM) & $\bigo{pn^{2}}$ \\
  SVGP & $\bigo{nm^{2} + m^{3} + dnm}$ \\
  KISS-GP & $\bigo{pn + pdm^{d} \log m}$ \\
  SKIP & $\bigo{drn + drm \log m + r^{3}n\log d + pr^{2}n}$\\
  \hline
\end{tabular}

%% file: sections/multi_task.tex
\section{APPLICATION 2: MULTI-TASK LEARNING}
\label{sec:multitask}
\begin{figure*}[t]
  \centering
  \includegraphics[width=\linewidth]{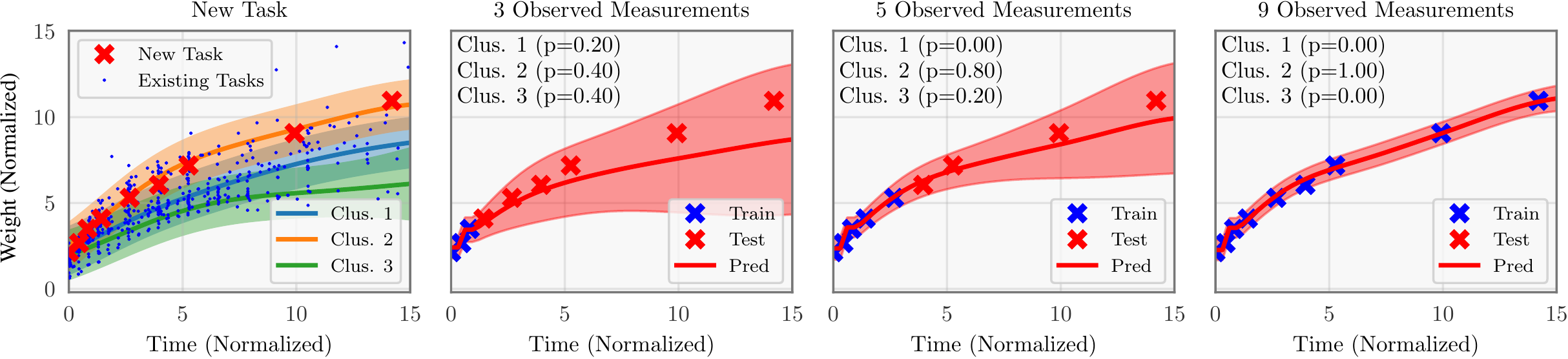}
  \caption{Applying the cluster-based MTGP model to new tasks. \label{fig:health_adaptation}}
  \vspace{-2ex}
\end{figure*}
\begin{figure}[t!]
  \centering
  \includegraphics[width=\columnwidth]{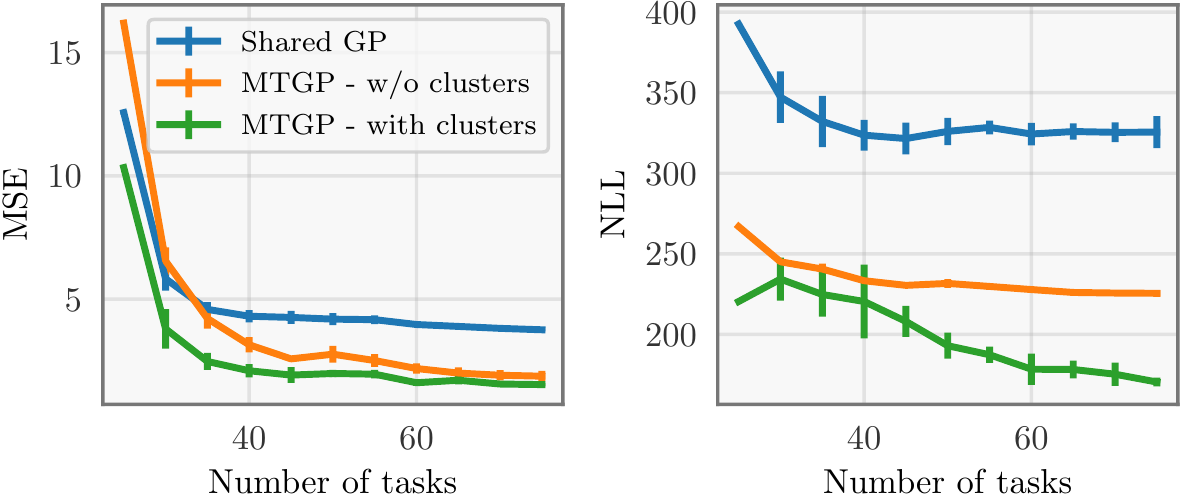}
  \caption{Predictive performance on childhood development dataset as a function of
    the number of tasks. \label{fig:health_comparison}}
  \vspace{-5ex}
\end{figure}
We demonstrate how the fast elementwise matrix vector products with SKIP can also be applied to accelerate multi-task Gaussian processes (MTGPs). Additionally, because SKIP provides cheap marginal likelihood computations, we extend standard MTGPs to construct an interpretable and robust multi-task GP model which discovers latent clusters among tasks using Gibbs' sampling.  We apply this model to a particularly consequential child development dataset from the Gates foundation.

\paragraph{Motivating problem.}
The Gates foundation has collected an aggregate longitudinal dataset of child development, from studies performed around the world.
We are interested in predicting the future development for a given child (as measured by
weight) using a limited number of existing measurements.
Children in the dataset have a varying number of measurements (ranging from 5 to 30), taken at irregular times throughout their development.
We therefore model this problem with a multitask approach, where we treat each child's development as a task.
This approach is the basis of several medical forecasting models \citep{alaa2017personalized,cheng2017sparse,xu2016bayesian}.

\paragraph{Multi-task learning with GPs.}
The common multi-task setup involves $\numpeople$ datasets corresponding to a set of different tasks, $\dset_{i}\!:\!\set{(\x^{(i)}_{1},y^{(i)}_{1}),...,(\x^{(i)}_{n_{i}}, y^{(i)}_{n_{i}})}_{i=1}^{\numpeople}$.
The multi-task Gaussian process (MTGP) of \citet{bonilla2008multi} extends standard GP regression to share information between several related tasks.
MTGPs assume that the covariance between data points factors as the product of kernels over (e.g. spatial or temporal) inputs and tasks.
Specifically, given data points $\x$ and $\x'$ from tasks $i$ and $j$, the MTGP kernel is given by
\begin{equation}
  k((\x, i),(\x', j)) = k_\text{input}(\x, \x)k_\text{task}(i, j),
  \label{eq:multitask_kernel}
\end{equation}
where $k_\text{input}$ is a kernel over inputs, and $k_\text{task}(i, j)$ -- the \emph{coregionalization kernel} -- is commonly parameterized by a low-rank covariance matrix $M = BB^\top \in \reals^{\numpeople \times \numpeople}$ that encodes pairwise correlations between all pairs of tasks.
The entries of $B$ are learned by maximizing \eqref{eq:marginalloglik}.
We can express the covariance matrix $K_\text{multi}$ for all $n$ measurements as
$$ K_\text{multi} = \K^{(\text{data})}_{\X\X} \circ \left( V BB^\top V \right), $$
where $V$ is an $n \times s$ matrix with one-hot rows: $V_{ij}=1$ if the $i^{th}$ observation belongs to task $j$.
We can apply SKIP to multi-task problems by using a SKI approximation of $K^{(\text{data})}$ and computing its Lanczos decomposition.
If $B$ is rank-$q$, with $q < n$, then we do not need to decompose $VBB^\top V^\top$ since the matrix affords $\bigo{n + sq}$ MVMs.\footnote{
  MVMs are $\bigo{n + sq}$ because $V$ has $\bigo{n}$ nonzero elements and $B$ is an $s \times q$ matrix.
}
For one-dimensional inputs, the time complexity of an MVM with $K_\text{multi}$ is $\bigo{n + m \log m + sq}$ -- a substantial improvement over standard inducing-point methods with MTGPs, which typically require at least $\bigo{nm^{2}q}$ time \citep{bonilla2008multi,alvarez2011computationally}.
For $n=4000$, SKIP speeds up marginal likelihood computations by a factor of $20$.
%


\paragraph{Learning clusters of tasks.}
Motivated by the work of \citet{rasmussen2002infinite}, \citet{shi2005hierarchical}, \citet{schulam2015framework}, \citet{hensman2015fast}, and \cite{xu2016bayesian}, we propose a modification to the standard MTGP framework. We hypothesize that similarities between tasks can be better expressed through $c$ latent subpopulations, or clusters, rather than through pairwise associations.
We place an independent uniform categorical prior over $\lambda_i \in [1, \ldots, c]$, the cluster assignment for task $i$.
Given measurements $\x_i, \x'_j$ for tasks $i$ and $j$, we propose a kernel consisting of product and sum structure that captures cluster-level trends and individual-level trends:
$$k(\x_i, \x'_j) = k_\text{cluster}(\x, \x') \delta_{\lambda_i = \lambda_j} + k_\text{indiv}(\x, \x') \delta_{i = j}.$$
Here, $k_\text{cluster}$ and $k_\text{indiv}$ are both Mat\'ern kernels ($\nu=\frac{5}{2}$) operating on $\x$, and the $\delta$ terms represent indicator functions.
Both terms can be easily expressed as product kernels.
We infer the posterior distribution of cluster assignments through Gibbs sampling.
Given $\lambda_{-i}$, the cluster assignments for all tasks except the $i^{th}$, we sample an assignment for the $i^{th}$ task from the marginal posterior distribution
\begin{equation*}
    p(\lambda_i | \y, \lambda_{-i}) \propto p(\y \mid \lambda_{-i}, \lambda_{i}=a \theta)p(\lambda_{-i},\lambda_{i}=a)
\end{equation*}
Drawing a sample for the full vector $\lambda$ requires $\bigo{cs}$ calculations of \eqref{eq:marginalloglik}, an operation made relatively inexpensive by applying SKIP to the underlying model.

\paragraph{Results.}
We compare the cluster-based MTGP against two baselines:
1) a single-task GP baseline, which treats all available data as a single task,
and 2) the standard MTGP. In \autoref{fig:health_comparison}, we measure the extrapolation accuracy for 25 children as additional children (tasks) are added to the model.
As the models are supplied with data from additional children, they are able to refine the extrapolations on all children.
The predictions of the cluster model slightly outperform the standard MTGP, and significantly outperform the single-task model.
Perhaps the key advantage of the clustering approach is interpretability: in \autoref{fig:health_adaptation} (left), we see three distinct development types: above-average, average, and below average.
We then demonstrate that as more data is observed when we apply the model to a new child with limited measurements, the model becomes increasingly certain that the child belongs to the above-average subpopulation.

%% file: sections/related_work.tex

%% file: sections/discussion.tex
\vspace{-1ex}
\section{DISCUSSION}
\vspace{-1ex}
\label{sec:discussion}

It is our hope that this work highlights a question of foundational importance for scalable GP inference: \emph{given the ability to compute $\A\bv$ and $\B\bv$ quickly for matrices $\A$ and $\B$, how do we compute $(A \circ B)\bv$ efficiently?}  We have shown an answer to this question can \emph{exponentially} improve the scalability and general applicability of MVM-based methods for fast Gaussian processes.

\paragraph{Stochastic diagonal estimation.} Our method relies primarily on quickly computing the diagonal in Equation \eqref{eq:two_mat_hpvm}. Techniques exist for stochastic diagonal estimation \citep{fitzsimons2016improved,hutchinson1990stochastic,selig2012improving,bekas2007estimator}. We found that these techniques converged slower than our method in practice, but they may be more appropriate for kernels with high rank structure.

\paragraph{Higher-order product kernels.}
A fundamental property of the Hadamard product is that $\textrm{rank}(A \circ B) \leq \textrm{rank}(A)\textrm{rank}(B)$
suggesting that we may need higher rank approximations with increasing dimension. In the limit, the SKI approximation $\W\K_{\U\U}\W^{\top}$ can be used in place of the Lanczos decomposition in equation \eqref{eq:two_mat_hpvm}, resulting in an exact algorithm with $\bigo{dnm + dm^{2}\log m}$ runtime: simply set $Q_{k}=\W$ and MVMs require $\bigo{n}$ time instead of $\bigo{nk}$, and set $T_{k}=\K_{\U\U}$ and MVMs now require $\bigo{m \log m}$ instead of $\bigo{k^{2}}$. This adaptation is rarely necessary, as the accuracy of MVMs with SKIP increases exponetially in $k$ in practice.

\paragraph{Space complexity.} To perform the matrix-vector multiplication algorithm described above, we must store the Lanczos decomposition of each component kernel matrix and intermediate matrices in the merge step for $\bigo{dkn}$ storage. This is better storage than the $\bigo{n^{2}}$ storage required in full GP regression, or $\bigo{nm}$ storage for standard inducing point methods, but worse than the linear storage requirements of SKI. In practice, we note that GPU memory is indeed often the major limitation of our method, as storing even $k=20$ or $k=30$ copies of a dataset in GPU memory can be expensive.